\documentclass[a4paper,twoside]{article}

\usepackage{epsfig}
\usepackage{subfigure}
\usepackage{calc}
\usepackage{amssymb}
\usepackage{amstext}
\usepackage{amsmath}
\usepackage{amsthm}
\usepackage{multicol}
\usepackage{pslatex}
\usepackage{apalike}
\usepackage{SCITEPRESS}     

\begin{document}

\title{Patch-based Carcinoma Detection on Confocal Laser Endomicroscopy Images - A Cross-Site Robustness Assessment}


\author{\authorname{Marc Aubreville\sup{1}, Miguel Goncalves\sup{2}, Christian Knipfer\sup{3,4}, Nicolai Oetter\sup{5,4}, Tobias W{\"u}rfl\sup{1}, Helmut Neumann\sup{6}, Florian Stelzle\sup{5,4}, Christopher Bohr\sup{2} and Andreas Maier\sup{1,4}} \affiliation{\sup{1}Pattern Recognition Lab, Computer Science, Friedrich-Alexander-Universit{\"a}t Erlangen-N{\"u}rnberg} \affiliation{\sup{2}Department of Otorhinolaryngology, Head and Neck Surgery, University Hospital Erlangen, Friedrich-Alexander-Universit{\"a}t Erlangen-N{\"u}rnberg} \affiliation{\sup{3}Department of Oral and Maxillofacial Surgery, University Medical Center Hamburg-Eppendorf} \affiliation{\sup{4}Erlangen Graduate School in Advanced Optical Technologies (SAOT), Friedrich-Alexander-Universit{\"a}t Erlangen-N{\"u}rnberg} \affiliation{\sup{5}Department of Oral and Maxillofacial Surgery, University Hospital Erlangen, Friedrich-Alexander- Universit{\"a}t Erlangen-N{\"u}rnberg} \affiliation{\sup{6}First Department of Internal Medicine, University Medical Center Mainz, Johannes Gutenberg-Universit{\"a}t Mainz}}


\abstract{Deep learning technologies such as convolutional neural networks (CNN) provide powerful methods for image recognition and have recently been employed in the field of automated carcinoma detection in confocal laser endomicroscopy (CLE) images. CLE is a (sub-)surface microscopic imaging technique that reaches magnifications of up to 1000x and is thus suitable for in vivo structural tissue analysis.\\
In this work, we aim to evaluate the prospects of a priorly developed deep learning-based algorithm targeted at the identification of oral squamous cell carcinoma with regard to its generalization to further anatomic locations of squamous cell carcinomas in the area of head and neck.
We applied the algorithm on images acquired from the vocal fold area of five patients with histologically verified squamous cell carcinoma and presumably healthy control images of the clinically normal contra-lateral vocal cord. 
\\
We find that the network trained on the oral cavity data reaches an accuracy of 89.45\% and an area-under-the-curve (AUC) value of 0.955, when applied on the vocal cords data. 
Compared to the state of the art, we achieve very similar results, yet with an algorithm that was trained on a completely disjunct data set. Concatenating both data sets yielded further improvements in cross-validation with an accuracy of 90.81\% and AUC of 0.970.
\\
In this study, for the first time to our knowledge, a deep learning mechanism for the identification of oral carcinomas using CLE Images could be applied to other disciplines in the area of head and neck. This study shows the prospect of the algorithmic approach to generalize well on other malignant entities of the head and neck, regardless of the anatomical location and furthermore in an examiner-independent manner.}

\onecolumn \maketitle \normalsize \vfill

\textbf{Erratum}: In earlier versions of this paper, the count of CLE image sequences for the vocal folds was inadequately reported to be 73. In reality, it was 47.  

\section{\uppercase{Introduction}}
\label{sec:introduction}

\noindent Squamous cell carcinoma is a common kind of cancer,  found in epithelial tissue. The prevalence within the head and neck region is estimated to be around 1.3 million cases per year \cite{Forastiere:2009bw,Ferlay:2014ht}.
 
Many cases of head and neck squamous cell carcinoma (HNSCC) are diagnosed at a late stage, which impairs treatment outcomes and increases mortality \cite{Muto:2004hy}. The gold standard of diagnosis is invasive biopsy of the tissue with subsequent histopathological assessment \cite{Oetter:2016cp}. However, biopsies carry the risk of infections and bleeding.  Furthermore, due to the invasiveness a limitation in the sample size and quantity hinders the finding of accurate resection margins \cite{Dittberner:2016jv,Nathan:2014ky}. An non- or minimally invasive in vivo characterization of microstructures could help detecting such malignancies at an early stage while at the same time reducing risk. Further, it could be of help for periodic monitoring of possibly malignant cellular structures, reducing the risk for unnecessary biopsies.

One method that has successfully been applied for visual inspection of suspicious lesions is Confocal Laser Endomicroscopy (CLE). In this imaging method, laser light is emitted and applied on tissue using a fibre-optic bundle that is typically inserted through the accessory channel of an endoscope \cite{chauhan2014confocal}. The resolution of CLE is high, providing magnifications of up to 1000x \cite{Oetter:2016cp} and enabling sub-cellular imaging. A contrast agent (fluorescein) is applied intravenously prior to the examination in order to stain the intercellular gap and hence outline the cell borders. CLE is successfully used in clinical routine diagnostics of the intestine \cite{Neumann:2010hb} and was recently also successfully applied on cancer assessment in the oral cavity \cite{Oetter:2016cp} and the upper respiratory tract \cite{Goncalves2017}.




Due to the property of making cellular structures visible, CLE is said to provide 'real-time' optical biopsies \cite{Parikh:2016gw}, which is a major advantage over the need to perform traditional biopsies, e.g. when finding a proper resection margin for intra-operative monitoring during surgical tumor removal. However, it was shown that the accuracy in interpretation of CLE images is highly dependent on the experience of the clinical expert, and that a significant learning curve exists \cite{Neumann:2011tc}. An automatic detection and interpretation of such images could thus help to improve the standard and make CLE also applicable with less training involved.

\begin{figure}
\centering
	\includegraphics[height=5cm]{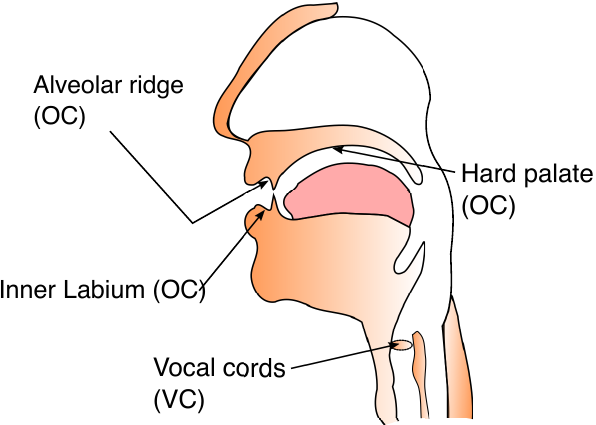}
	\caption{Anatomical locations from the oral cavity and the upper aero-digestive and respiratory tract.  }
	\label{locations}
\end{figure}

Deep learning methods, such as convolutional neural networks (CNN) have recently been used in a variety of image recognition tasks. We have shown that CNN-based recognition methods outperform the state of the art in HNSCC detection on CLE images, using a data set of 12 patients \cite{DBLP:journals/corr/AubrevilleKOJRD17}. In order to investigate the robustness of the method, generalization has to be assessed to other environments. One step into showing this generalization is to apply a trained machine learning model from one anatomical site and clinical team to another, without any modification of the underlying model structure and content. This would provide a strong hint of generalization towards other locations of the upper aero-digestive tract with similar but not identical epithelia.

\section{\uppercase{Material}}

For the present work, we are using images from two anatomical locations (see figure  \ref{locations}):  From within the oral cavity, we used images from three clinically normal sites and a lesion site with verified SCC. From the upper aero-digestive and respiratory tract, we used images of the vocal cords (clinically normal and with verified malign changes). All images were acquired using a probe-based CLE (pCLE) system (Cellvizio, Mauna Kea Technologies, Paris, France). From all patients, written informed consent was obtained prior to the study. Approval was granted by the respective institutional review boards. The research was carried out in accordance with the Code of Ethics of the World Medical Association (Declaration of Helsinki) and the guidelines of the Friedrich-Alexander University Erlangen-Nuremberg.

\begin{figure*}
	
\centering
	\includegraphics[height=4cm]{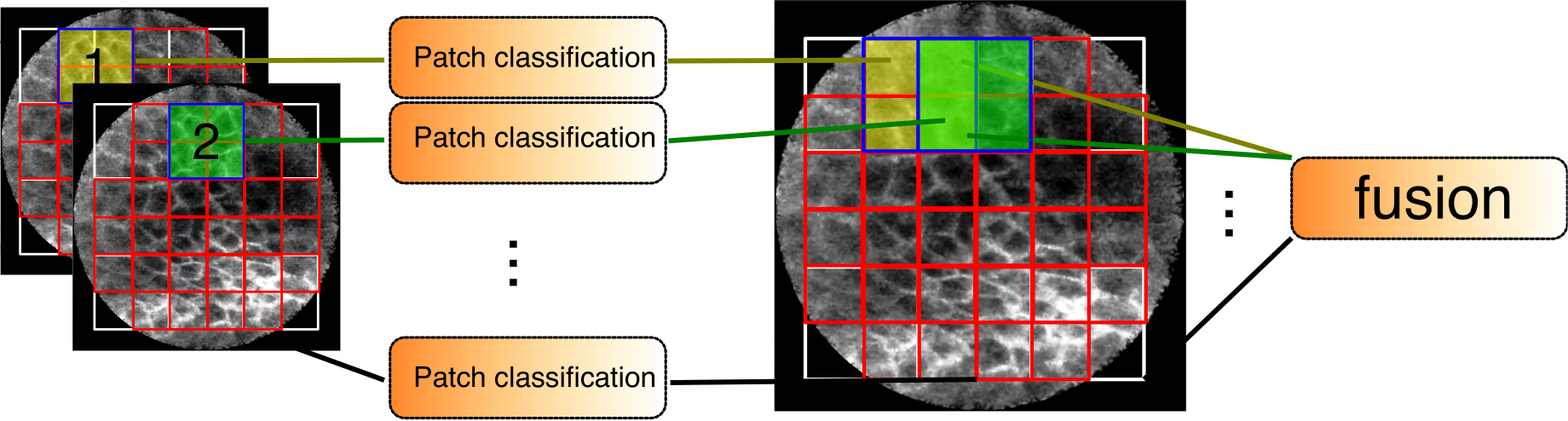}
	\caption{Overview of the classification method. Patches (1,2,...) are extracted from the visible view area of the image and independently fed into a CNN-based patch classification. The resulting probabilities are then fused in a final step to one class-probability tuple for the whole image.}\label{ppf}
\end{figure*}

\subsection{Oral Cavity (OC)}

We included image sequences ($N=116$) from 12 patients with diagnosed and verified HNSCC in the oral cavity that were recorded at the Department of Oral and Maxillofacial Surgery (University Hospital Erlangen)\footnote{Study approved by the ethics committee of the University of Erlangen-N{\"u}rnberg; reference number: 243\_12 B}. 
For all patients, imaging was performed at the suspected carcinoma site, as well as three other anatomical sites within the oral cavity. After verification of the carcinoma diagnosis by histo-pathology, all patients underwent surgery for removal of the suspected tissue. Oetter \textit{et al.} found that the accuracy as rated by CLE-experienced specialists was 92.3\%, where the experts were allowed to see the complete video sequence to base their assessment upon. So, this number accounts for occurrence of singular or sparse cues for a correct classification. It is to be expected that an assessment of a whole sequence can achieve a better performance than on singular images. In contrast, we performed evaluation on a per-frame base. The total number of images with sufficient quality in this data set is 7,894.

\subsection{Vocal Cords (VC)}

Squamous cell carcinoma of the vocal cords is the most prevalent form of cancer of the laryngeal tract \cite{Parkin:2005ig} and an anatomy where bright light endoscopy alone is insufficient for diagnosis, because a significant range of benign mucosal alterations with similar macroscopic appearance exist \cite{Goncalves2017}. Biopsies require a sufficient amount of material and extensive biopsy in this anatomical region causes functional problems, such as chronic hoarseness or other voice modifications \cite{Cikojevic:2008jd}. Accurate diagnosis using optical, non-invasive methods would thus provide a significant improvement for the life quality of patients with suspected cancer of the upper respiratory tract. 

In this study, we included $N=47$ image sequences from five patients with verified malignant transformations of the vocal folds. For each patient, the contra-lateral vocal cord (that was clinically normal in all cases) was also investigated using CLE, building a presumably healthy data set. The images were acquired at the Department of Otorhinolaryngology, Head and Neck Surgery, University Hospital Erlangen\footnote{Study approved by the ethics committee of the University of Erlangen-N{\"u}rnberg; reference number: 60\_14 B}. All image sequences were taken during micro-laryngoscopy and were pre-selected by a clinical expert in CLE imaging, in order to remove images where the acquisition was not performed properly, e.g. when the probe was not in contact with the tissue. For one patient only image sequences of the clinically normal vocal cord exist in a sufficient quality, reducing the number of carcinoma patients to four. The total number of images in this data set is 4,425. Previous studies of automatic carcinoma detection on this data set have shown accuracies in grading of between 86.4\% and  89.8\% \cite{Vo2016}, using  the methods of Jaremenko \cite{Jaremenko:2015kh}. 

Goncalves \textit{et al.} selected 31 representative images out of a different vocal cord data set of 7 patients and found rating accuracies by ENT specialists to be between 58.1\% and 87.1\%, where the non-CLE experienced doctors had a mean accuracy of 67.7\% and those with profound experience of CLE one of 82.2\% \cite{Goncalves2017}. Also for this data set, the differences in recognition performance are well explained by the learning effect as described by Neumann \cite{Neumann:2011tc}. However, the relatively low accuracies for less experienced observers also motivate the development of an automated approach, as pointed out by Goncalves. 



\begin{figure*}
\centering
	\includegraphics[height=7cm]{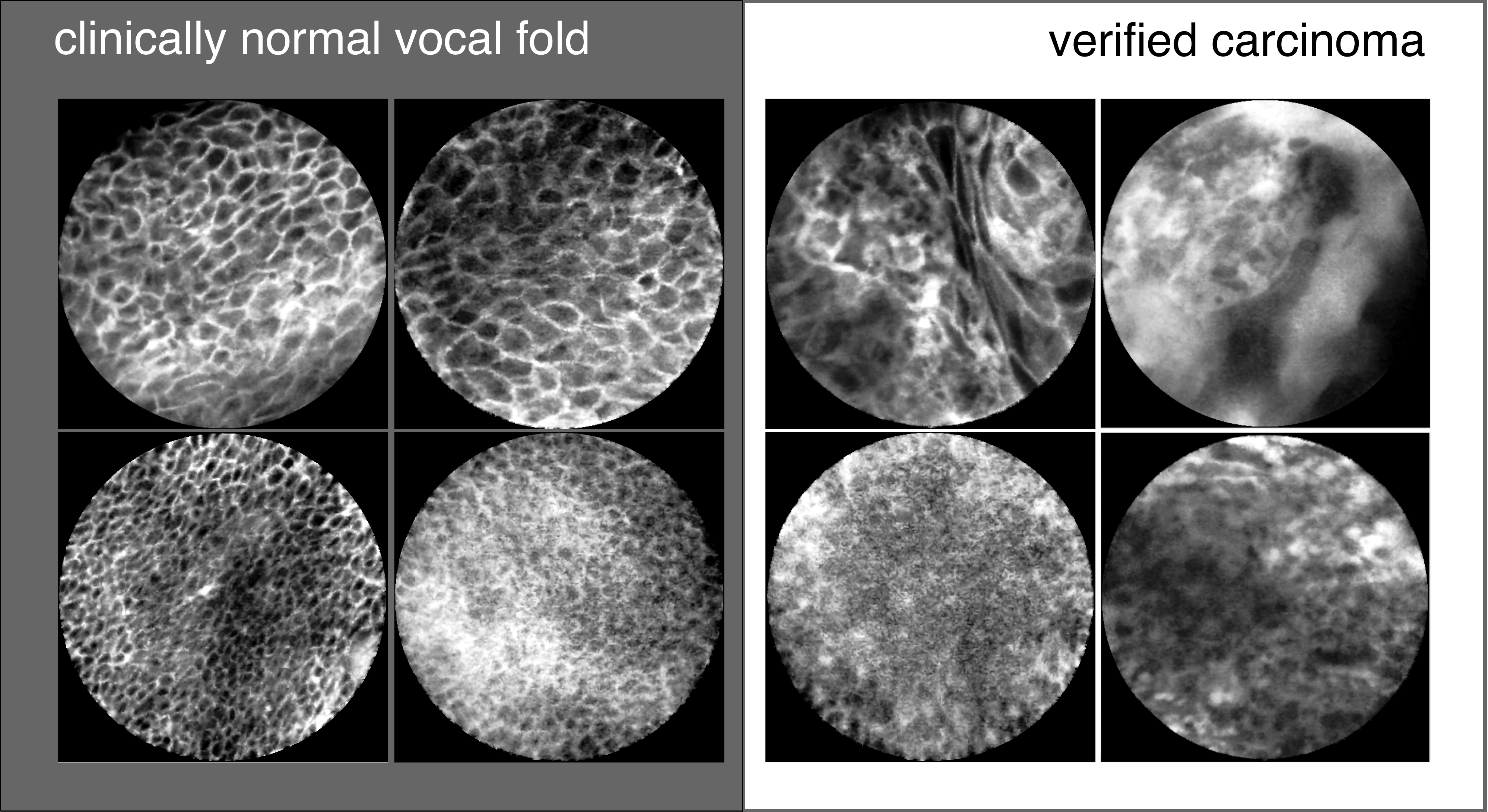}
	\caption{Selected CLE Images acquired from the vocal fold area. On the left, presumably healthy images are depicted that were taken from the clinically normal contralateral vocal cord of patients with epithelial cancer one of the both vocal cord. }
	\label{CLE_VC_data}
\end{figure*}

\begin{figure*}
\centering
	\includegraphics[height=7cm]{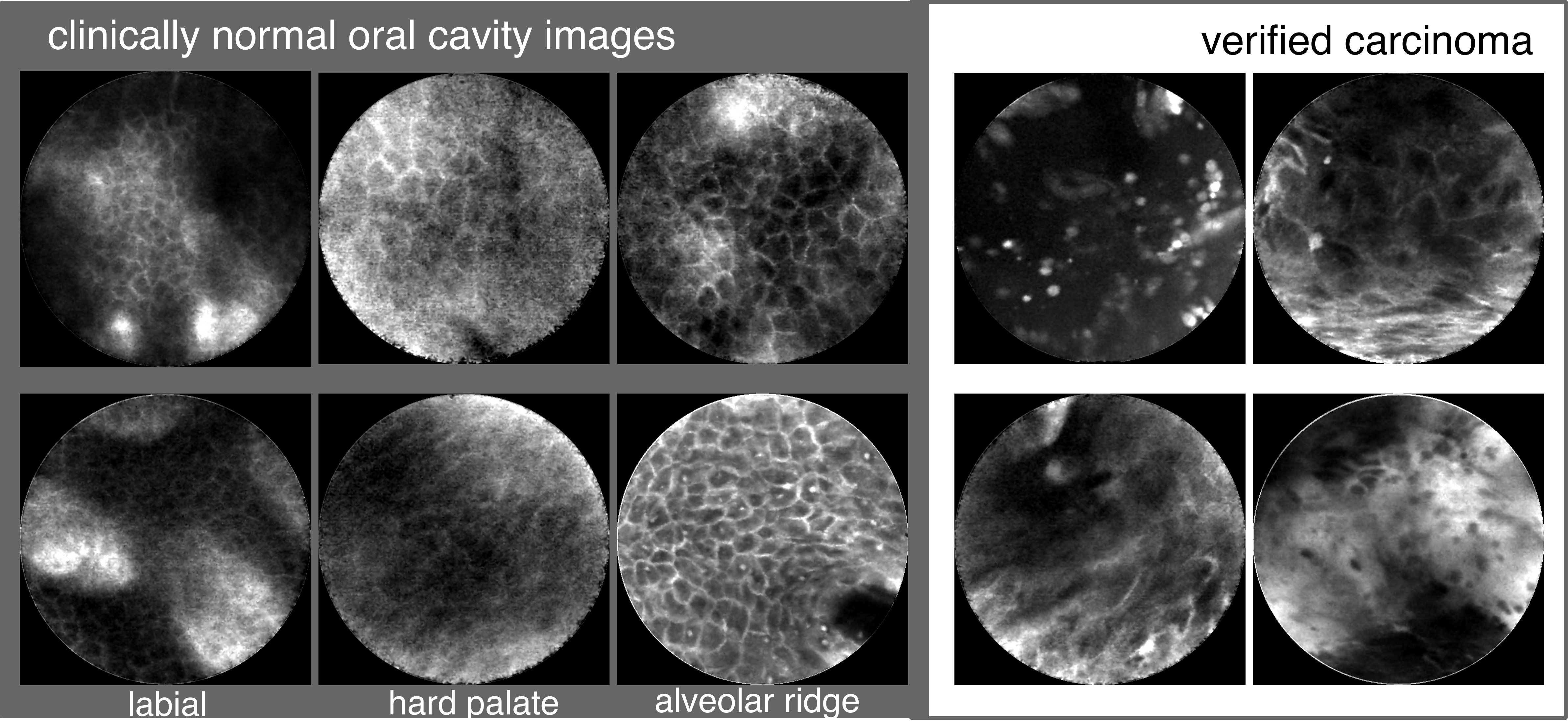}
	\caption{Selected CLE Images acquired from the oral cavity. On the left, presumably healthy images are depicted that were taken from clinically normal regions from patients suffered from histologically verified head and neck squamous cell carcinoma.}
	\label{CLE_OC_data}
\end{figure*}

\subsection{Image Quality and Artifact Occurrence}

CLE images suffer from a range of deteriorations in image quality \cite{Neumann:2012ic,DBLP:journals/corr/AubrevilleKOJRD17}. While some images are tainted by motion artifacts, others have a very low signal to noise ratio (see figure \ref{CLE_OC_data}, second column).

\begin{table*}[ht!]
	
\centering
\begin{tabular}{|l|c|c|c|c|c|}
\hline
 Name & 	Train data set	&  Patients training &  Test data set & Patients test & Cross validation \\
 \hline
OC & Oral cavity & 11  & Oral cavity & 1 & LOPO \\
VC & Vocal cords & 4 & Vocal cords & 1 & LOPO \\
OC/VC & Oral cavity & 12 & Vocal cords & 5 & - \\
VC/OC & Vocal cords & 5 & Oral cavity & 12 & -\\
OC+VC & both & 16 & both & 1 & LOPO \\
\hline
\end{tabular}

\caption{Test conditions. For cross-correlation, a leave-one-patient out scheme was used. }
\label{table}
\end{table*}

\begin{figure}
\centering
\includegraphics[width=0.5\textwidth]{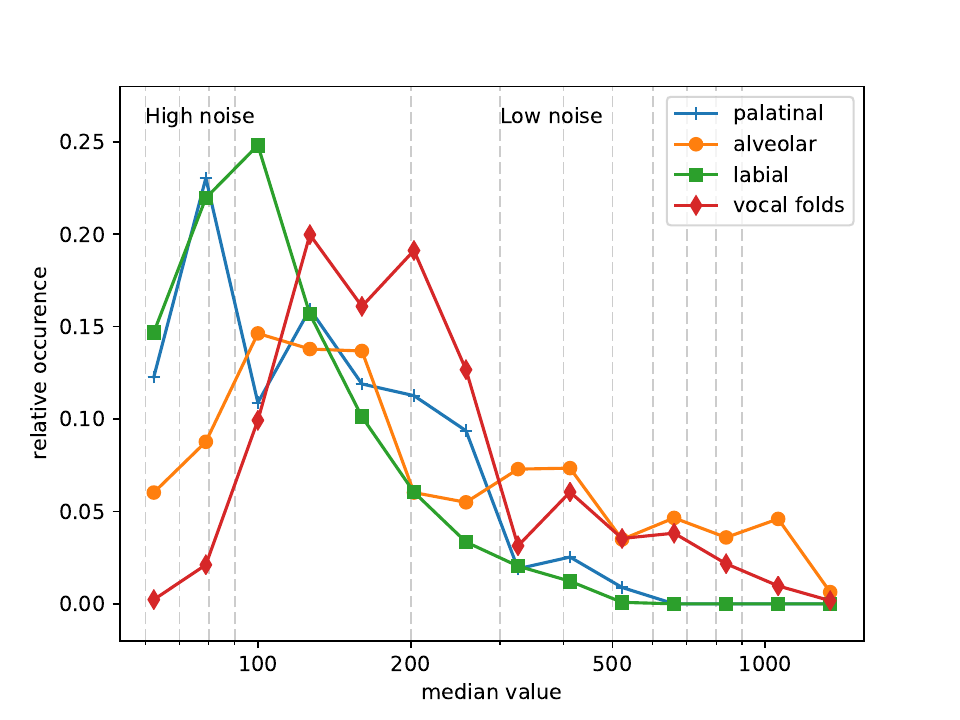}
\caption{Normalized histogram of the median value for the different classes of clinically normal tissue from both data sets. Due to the wide range of pixel values, the histogram is given at log scale. }
\label{median}	
\end{figure}

For images acquired by CLE, image noise intensity is mostly dependent on the actual amount of received photons, i.e. the optical response of the tissue, since the optical receiver has a broad dynamic range and images are automatically range compressed by the viewing software to fit the gray level range (8bit) of today's screens and image file formats. One metric to measure this optical response is by the pixel value distribution of the image. We calculated the median pixel value of each image to assess the optical response and compared the different anatomical locations of clinically normal tissue (see figure \ref{median}). CLE images have sparse occurrences of vessels (see \ref{CLE_OC_data}, top left), which are typically very bright, which is why the median might be beneficial for contrast assessment of the cellular structures of the overall image. Low values of the median indicate a noisy image, while high values indicate an image with a high optical response to the laser. For low values, the range compression of the viewing system is in fact an extension, which amplifies noise further for the examiner.

We find that a significant amount of images of the palatinal and labial region have a low response, indicating high noise levels, which is also confirmed by optical inspection of the images. In contrast, images of the alveolar ridge and - even more - the vocal folds tend to have a better signal to noise ratio, as depicted in figure \ref{median}.

This can be related to the different anatomical properties of the respective epithelia: Regions with high mechanical stress due to chewing have a higher degree of cornification \cite{Rohen:2000ux}.
Specifically, the hard palate is known to have a high degree of cornification \cite{rauch2015taschenbuch}. The inner lip (labium) is generally not considered a cornified epithelium, however our images were taken at the intersection between mucous membrane and outer lip with its epidermal layer, where cornification is indeed prominent \cite{Rohen:1994tv}. 
Contrary to this, the vocal cords are known to consist of multiple layers of uncornified squamous epithelium \cite{Rohen:1994tv}.

The difference in image quality could, however, also be caused by a  different preselection bias between the two clinical teams.

In our data sets, motion artifact incidence is significantly higher in the oral cavity images compared to the vocal cords images. This can be related to patients being under general anaesthesia in case of the vocal fold microendoscopy \cite{Goncalves2017}. In this case, the only reason for motion is hand movement of the clinician performing the image acquisition. 

\section{\uppercase{Methods}}

Our method is based on the extraction of squared patches from the round field of view area of a CLE image, classification by a deep convolutional network (CNN). Subsequently, the a posteriori probabilities are fused (see Fig. \ref{ppf}; for more details, see \cite{DBLP:journals/corr/AubrevilleKOJRD17}). The approach limits overfitting of the CNN model by a small patch size (80x80 px) and thus a reduced capacity of the network. Additionally, the strategy results in a large number of training samples, since every image consists of a multitude of patches. We trained the network for 60 epochs, using the Adam optimizer at an initial step size of 0.01 within the TensorFlow framework. 

In total, we performed two additional test sets: 
\begin{enumerate}
	\item \subsubsection*{Generalization tests} We performed training of our deep convolutional models on one anatomical location and testing on the other (tests OC/VC and VC/OC, see table \ref{table}). 
	\item \subsubsection*{Algorithmic validation} We performed a validation of the automatic detection algorithm \cite{DBLP:journals/corr/AubrevilleKOJRD17} on the vocal cords data set and on the concatenated data set (tests VC and OC+VC, respectively).
\end{enumerate}


For all tests, where train and test data were taken from the same data set, we applied leave-one-patient-out cross validation. Independent x-fold cross-validation or simple random train-test-splitting isn't applicable, since high correlations between consecutive frames within one sequence might exist. The VC data set is small compared to the other data sets, has a much lower number of patients and comes from a small anatomical structure. This leads us to expect generalization to the oral cavity to work better than vice versa. 

Our intention for the last test (OC+VC) is, how well the algorithm is able to improve overall results from more image material. 

\section{\uppercase{Results}}
\begin{figure}
\centering
\includegraphics[width=0.50\textwidth]{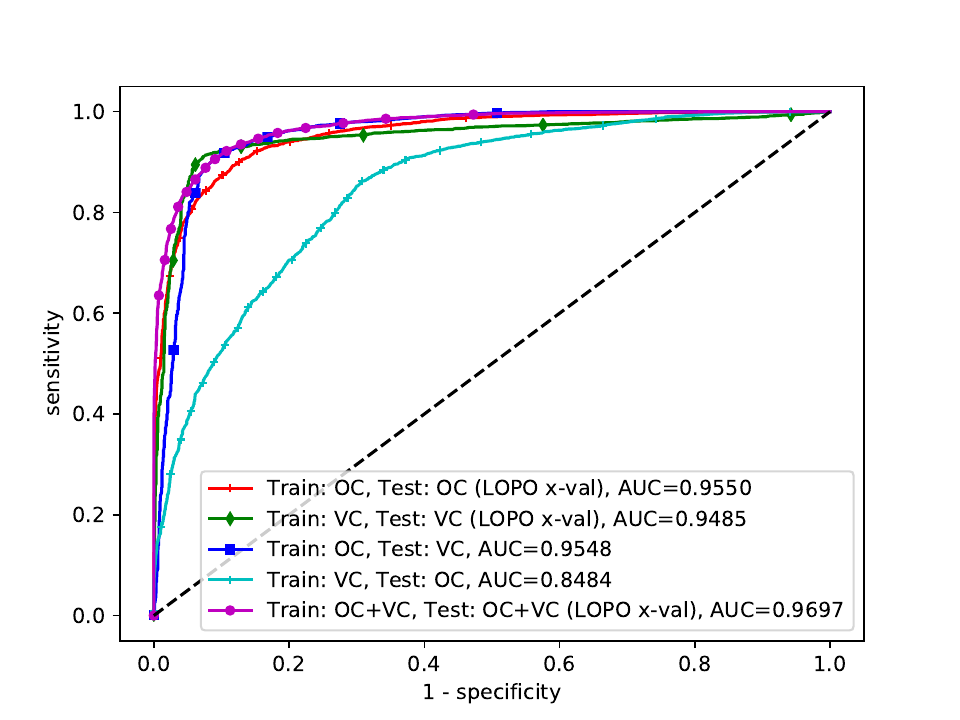}	
\caption{Receiver Operating Characteristic (ROC) curve for the different setups. (OC=oral cavity, VC=vocal cords)}
\label{roc}
\end{figure}

\begin{table*}[ht!]
\centering
	\begin{tabular}{|l|c|c|c|c|}
	\hline
		Condition & Accuracy & Precision & Recall & ROC area under curve \\
		\hline
		OC \cite{DBLP:journals/corr/AubrevilleKOJRD17} & 88.34\% & 85.40\%   & 91.10\% & 0.9550 \\
		VC & 91.39\%  & 93.64\% & 92.03\% & 0.9484\\
		OC/VC & 89.45\%  & 87.47\% & 96.37\% & 0.9548\\
		VC/OC & 68.53\%  & 60.81\% & 95.63\% & 0.8484 \\
		OC+VC & 90.81\%  & 90.12\% & 92.59\% & 0.9697
\\ 
		\hline
	\end{tabular}
	\caption{Results of all tests. For the cross validation cases OC, VC and OC+VC, the results were calculated on the concatenated result vector of all cross validation steps.}
	\label{results}
\end{table*}

\begin{figure}
\centering
\includegraphics[width=0.5\textwidth]{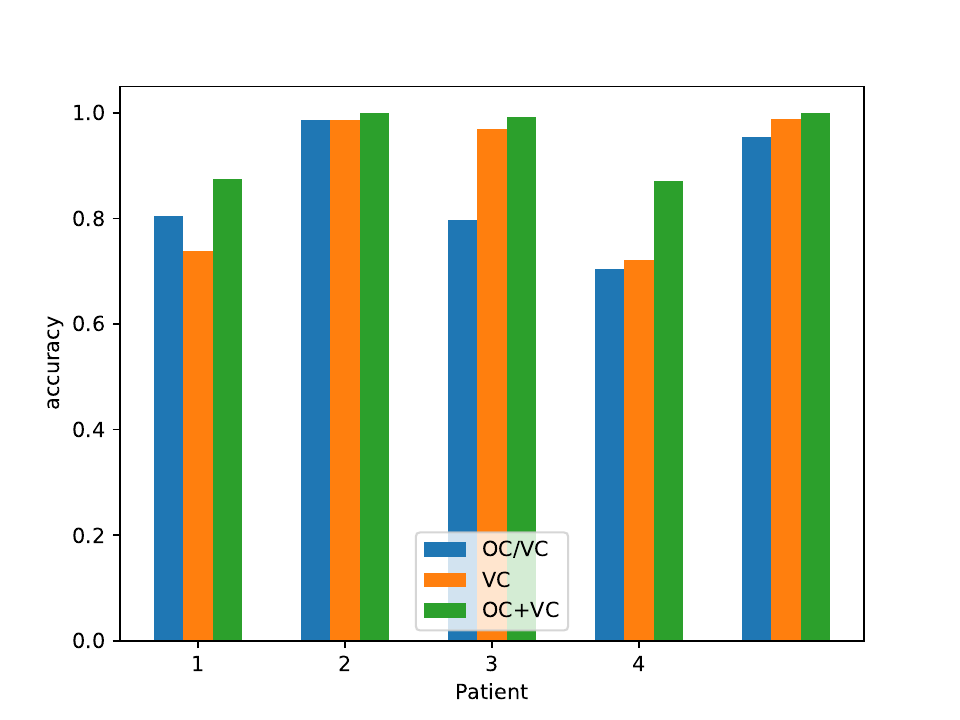}
\caption{Accuracy for all patients with both classes of the vocal fold data set.}	
\label{auc_vc}
\end{figure}

We find that the patch-based classification method seems to generalize well from the oral cavity data set to the vocal cords dataset (ROC area-under-the-curve of 0.9548). It is a comparable figure to the original data, where the ROC AUC was 0.9550. Trained on the vocal cords data set, the method outperforms the cross-validation results as reported by Vo \textit{et al.} for the method by Jaremenko \textit{et al.} slightly \cite{Vo2016,Jaremenko:2015kh}. 
When comparing the results on individual patients, slight differences between both approaches occur, where the approach only trained on VC data performs better for patient 3, while the approach trained on OC data performs better for patient 1. The concatenated data set increased performance for all patients in cross-validation (cf. figure \ref{auc_vc}).

The generalization task from the vocal cord data set to the much larger oral cavity data set, however, did not show comparable results, having AUC values of only 0.8484. Inspecting individual patient performance, it is obvious that the generalization loss is prominent in a number of patients, while others, like the tests on patient 4,5,6 and 10 perform comparable to the tests on the original OC data set.

When the data set is concatenated (condition OC+VC), the accuracy and ROC AUC values increases, with values of 90.81\% and 0.9697, respectively.

\begin{figure}
\centering
\includegraphics[width=0.5\textwidth]{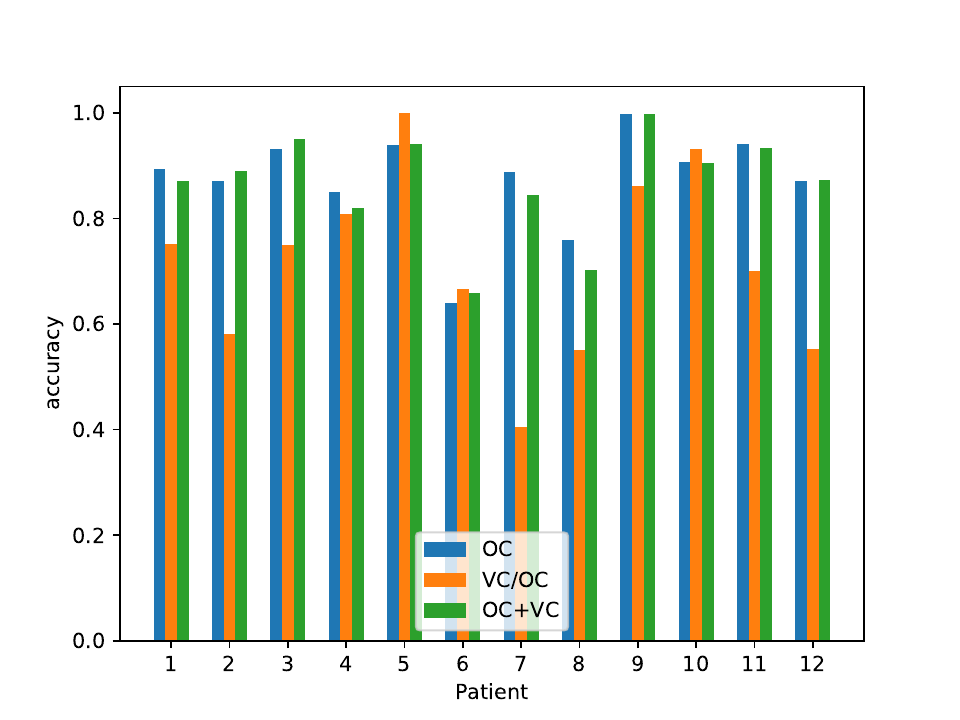}
\caption{Accuracy for all patients of the oral cavity data set.}	
\label{auc_oc}
\end{figure}

\section{\uppercase{Discussion}}

The much greater variance of the oral cavity data set (cf. figure \ref{median}) due to the larger variety in acquisition conditions led to better generalization properties compared to the vocal cords data set. The general signal-to-noise ratio was much better in this case. This is the reason why the classifier trained on the vocal cords tends to confuse noisier images, as they have been recorded from the cornified sections of the oral cavity, for malignant tissue. This is also reflected by the high recall and low precision ratings in this classification task (cf. 4th row of table \ref{results}). 

For the generalization from the oral cavity to the vocal fold data set, this restriction did not apply, since the CLE imaging conditions within the oral cavity seems to be a superset of those on the vocal folds.  However, also the greater number of patients for whom verified carcinoma imaging material was available likely played a role, which is also indicated by the increased performance in cross-validation for the concatenated data set. This indicates that the pattern recognition capacity of the model is not yet reached and that additional imaging data would likely increase performance further.

Since histological verification was only present for cancerous areas in both data sets, we can only assume that clinically normal regions represent healthy epithelium. Extraction of tissue from those regions would however be ethically questionable and not receive approval of the review boards.

It is questionable, if an 100\% accurate classification of epithelial tissue is possible using CLE alone, as even experts in the field of CLE were not able to classify cancerous tissue perfectly \cite{Goncalves2017,Oetter:2016cp}. Due to the low penetration depth of CLE, it is sometimes possible to overlook tumors that spread within the submucosa. Such tumors could be visualized only through histological section or perhaps through Optical Coherence Tomography \cite{Betz:2015ge}. 

One important aspect in automated inspection of CLE images is the removal of artifact-tainted images prior to training, since artifact occurrence is correlated with the surface conditions of the epithelium. This implies that it is also correlated to the malignancy classification, a causal relationship between artifact prevalence and tissue classification should, however, be neglected. This step was done manually in this work and this problem is subject of future work.

Even though our approach found good accuracy ratings, generalization can not be claimed to be fully shown with this study, due to the limited amount of patient data. Because of this, future work of our research group will concentrate on the acquisition of imaging data in order to increase the variance in the data set, which will presumably increase performance and robustness of the algorithmic approach.

\section{\uppercase{Summary}}
In this work, we have shown the principal ability to generalize patch-based CLE image classification with convolutional networks of potentially cancerous epithelium from a more diversified data set (from the oral cavity) to one of another anatomical location (the vocal folds) with less variance. The second data set was from a different clinic and a different team.

The generalization showed very promising results and concatenation of both sets did show further improvements in a leave-one-patient-out cross-validation scenario.

In total, we achieved an accuracy of $89.45\%$ in the generalization task, where the classification model was trained on the oral cavity data set and applied on the vocal cords data set. For the concatenated data set with 17 patients, we achieved a total accuracy of $90.81\%$ for the complete data set.  

%

\bibliographystyle{apalike}
{\small

\vfill
\end{document}